\documentclass[article, 11pt]{article}

\usepackage{arxiv}
\usepackage[numbers]{natbib}

\usepackage{microtype}
\usepackage{graphicx}
\usepackage{subfigure}
\usepackage{booktabs} %
\usepackage{hyperref}

\usepackage{amsmath}
\usepackage{amssymb}
\usepackage{mathtools}
\usepackage{amsthm}
\usepackage{algorithm}
\usepackage{algorithmic}

\usepackage[capitalize,noabbrev]{cleveref}

\usepackage[textsize=tiny]{todonotes}

\usepackage{parskip}
\usepackage{graphicx}
\usepackage[T1]{fontenc} 
\usepackage{epsf} 
\usepackage{graphics} 
\usepackage{color,etoolbox}
\usepackage{bbm}
\usepackage{mysymbols}
\usepackage{comment} 
\usepackage{pifont}
\usepackage{subfigure}
\usepackage{multirow}
\usepackage{multicol}
\usepackage{wrapfig}
\usepackage{color, colortbl}
\usepackage{float}

\title{Representer Point Selection for Explaining Regularized High-dimensional Models}
\author{
    {Che-Ping Tsai}\\
    Carnegie Mellon University\\
    Pittsburgh, PA 15213\\
    \texttt{chepingt@andrew.cmu.edu}
    \And
    {Jiong Zhang}\\
    Amazon Search\\
    Palo Alto, CA 94301\\
    \texttt{zhangjiong724@gmail.com}
    \And
    {Eli Chien}\\
    University of Illinois Urbana-Champaign\\
    Champaign, IL 61801 \\
    \texttt{ichien3@illinois.edu}
    \And
    {Hsiang-Fu Yu}\\
    Amazon Search\\
    Palo Alto, CA 94301\\
    \texttt{rofu.yu@gmail.com}
    \And
    {Cho-Jui Hsieh}\\
    University of California, Los Angeles\\
    Los Angeles, CA 95005\\
    \texttt{chohsieh@cs.ucla.edu}
    \And
    {Pradeep Ravikumar}\\
    Carnegie Mellon University\\
    Pittsburgh, PA 15213\\
    \texttt{pradeepr@andrew.cmu.edu}
}

\begin{document}

\date{}
\maketitle

\begin{abstract}
We introduce a novel class of sample-based explanations we term \emph{high-dimensional representers}, that can be used to explain the predictions of a regularized high-dimensional model in terms of importance weights for each of the training samples. Our workhorse is a novel representer theorem for general regularized high-dimensional models, which 
decomposes the model prediction in terms of contributions from each of the training samples: with positive (negative) values corresponding to positive (negative) impact training samples to the model's prediction. 
We derive consequences for the canonical instances of $\ell_1$ regularized sparse models, and nuclear norm regularized low-rank models. 
As a case study, we further investigate the application of low-rank models in the context of collaborative filtering, where we instantiate high-dimensional representers for specific popular classes of models. Finally, we study the empirical performance of our proposed methods on three real-world binary classification datasets and two recommender system datasets. We also showcase the utility of high-dimensional representers in explaining model recommendations. 
\end{abstract}

\section{Introduction}
\label{sec:intro}

Sample-based explanations aim to explain a machine learning model's prediction by identifying the most influential training samples that led to the prediction. This is usually done by measuring the influence of each training sample on the model's prediction scores. The explanations not only assist users in understanding the rationale behind the prediction, but also allow  model designers to debug or de-bias the training data~\citep{kong2021resolving,thimonier2022tracinad}. 

To measure the impact of each training sample on the prediction score, a classical technique is to compute the derivative of the prediction score with respect to each training instance using implicit function theory, an approach also known as influence functions~\citep{cook1980characterizations,koh2017understanding}. However, computing the  influence function requires the inversion of the Hessian matrix, causing significant scalability issues when handling large models. 
To compute sample-based explanations in an efficient manner, another  method called {\bf Representer Point Selection} has been developed~\cite{yeh2018representer}. 
This method is based on the classical representer theorem~\citep{scholkopf2001generalized}, which states that a regularized empirical risk minimizer over a reproducing kernel Hilbert space (RKHS) can be decomposed into a linear combination of kernel functions evaluated on each training sample. While functions parameterized with neural networks do not necessarily lie in a pre-specified RKHS, \citet{yeh2018representer} propose to treat the last layer of a neural network as a linear machine and the remaining part as a fixed feature encoder. Upon fine-tuning the last layer with $\ell_2$ regularization, the representer theorem can then be applied, allowing us to obtain importance scores of the training data. In this development, the use of $\ell_2$ regularization served as a RKHS norm with respect to linear kernels, which was key to recruiting the representer theorem. 

However, $\ell_2$ regularizers are not always suitable for high-dimensional models where the number of parameters might even be larger than the number of samples, and where the model parameters might lie in a lower dimensional sub-space. 
In such settings, in order for the resulting estimators to have strong statistical guarantees, it is often critical to employ high-dimensional regularizations that encourage the model parameter to lie in such lower-dimensional structured subspaces~\citep{negahban2011estimation}. Two canonical instances of such high-dimensional regularizers include the $\ell_1$ norm regularization that encourages parameter vectors to have sparse structure, and the nuclear norm regularization imposes low-rank structure on parameter matrices. The caveat however is that these regularizations cannot typically be cast as RKHS norms, and thus the classical representer theorem does not apply. Therefore, it remains unclear how to select representer points for high-dimensional models, despite the widespread use of high-dimensional models in practical applications such as compressed sensing~\citep{donoho2006compressed} and recommender systems~\citep{candes2008exact,recht2011simpler}.

We first present a general theorem that provides a representer theorem for regularized high-dimensional models, where we leverage the rich structure of the regularization sub-differentials, as well as the analytical framework of \citet{negahban2012unified} that associates the regularization functions with a collection of structured low-dimensional subspaces. We term the resulting sample-based explanations for these high-dimensional models as \emph{high-dimensional representers}. As with the original representer points for $\ell_2$ regularized models, there is a global importance score per training sample, as well as a local importance score that measures the similarity between the test point and the training sample. But unlike the $\ell_2$ regularized case, the representer theorem entails that this local similarity is measured after an appropriate linear projection of the test input and the training sample to the structured model parameter subspace. Thus, even in cases where the model parameters might be quite high-dimensional, the local similarity is quite meaningful, as well as scalable and efficient since it is computed over a much lower dimensional structured subspace.

Given the general theorem, 
 we then derive its consequences for the important settings of sparse vectors with $\ell_1$ regularization, and low-rank matrices with nuclear norm regularization, leading to sample-based explanation methods under those high-dimensional regularizers.
Equipped with the results, 
we explore the use of our technique in the context of collaborative filtering, including various specific model instances such as collaborative matrix factorization models~\citep{koren2009matrix}.  We also investigate deep neural network variations of these models, the two-tower models~\citep{mao2021simplex, li2019multi}, by treating the final interaction layer is treated as a bilinear matrix factorization model and the other layers are fixed encoders when applying our method. This cannot be done with the $\ell_2$ representer methods as the final layer is a product of two matrices. 
Lastly, we evaluate the empirical performance of the high-dimensional representers on three real-world binary classification datasets and two recommender system datasets. We also demonstrate the practical utility of high-dimensional representers in explaining the recommendations generated by our models.

\section{Related Work}
\label{sec:related}
Prominent approaches for estimating training data influence to a test point include influence functions~\citep{koh2017understanding}, representer point selection~\citep{yeh2018representer}, and TracIn~\citep{pruthi2020estimating}. Influence functions~\citep{wojnowicz2016influence, khanna2019interpreting,bae2022if} estimate training sample importance by measuring "how the model's prediction change if we remove a particular training sample and retrain the model."
However, computing influence functions requires calculating the inverse of the Hessian matrix. Exact estimation requires time complexity at least quadratic to the number of parameters and is thus unsuitable for large or high-dimensional models~\citep{guo2020fastif,hammoudeh2022identifying,schioppa2022scaling}. 

TracIn quantifies training data importance by measuring similarities between gradient at training and test samples over trajectories~\citep{yeh2022first,chen2021hydra}.
However, their approach only applies to models trained with stochastic gradient descent, which may not be an efficient way for high-dimensional model training.
Also, TracIn requires storing and accessing checkpoints of models during training and is not applicable to off-the-shelf models. 
The most relevant work to ours is the ($\ell_2$) representer point selection: \citet{brophy2022adapting} extends it to explain decision trees using supervised tree kernels. \citet{sui2021representer} improves it with local Jacobian expansion. Another line of sample-based explanations relies on repeated retraining ~\citep{ghorbani2019data,jia2019efficient,kwon2021beta,feldman2020neural}, which are more costly compared to the methods mentioned above since it
requires retraining models multiple times.

On the other hand, representer theorems~\citep{scholkopf2001generalized} in machine learning have targeted non-parametric regression in RKHS. \citet{bohn2019representer} connect representer theorems and composition of kernels. \citet{unser2019representer} derive general representer theorems for deep neural networks and make a connection with deep spline estimation. \citet{unser2016representer} also propose representer theorems for $\ell_1$ regularization, but their theorems have a different formulation for a difference purpose: they attribute model parameters to basis on the nonzero coordinates to show that the minimizer is sparse. In our work, we consider a simpler task of explaining regularized high-dimensional models and develop novel representer theorems for this purpose.

\section{Preliminary}
\label{sec:prelim}
Before providing our general framework for high-dimensional representers, it is instructive to recall classical machinery in high-dimensional estimation. As \citet{negahban2012unified} show, we can think of structure in high-dimensional models as being specified by collections of lower-dimensional subspaces. 

\textbf{Example: Sparse Vectors:} Consider the set of $s$-sparse vectors in $p$ dimensions. For any particular subset $S \subseteq \{1,\hdots,p\}$, with cardinality $s$, define the subspace: $A(S) = \{\theta \in \mathbb{R}^p\,:\,\theta_j = 0,\quad \forall j \not\in S\}$.
It can then be seen an $s$-sparse vector lies in one of the collection of low-dimensional subspaces $\{A(S)\}_{S \subseteq [p]}$.

\textbf{Example: Low-Rank Matrices:} For any matrix $\Theta \in \mathbb{R}^{d_1 \times d_2}$, let $\text{col}(\Theta) \in \mathbb{R}^{d_1}$ its column space, and $\text{row}(\Theta) \in \mathbb{R}^{d_2}$ denote its row space. For a given pair $(U,V)$ or $k$-dimensional subspaces $U \subseteq \mathbb{R}^{d_1}$ and $V \subseteq \mathbb{R}^{d_2}$, we can define the subspaces:
$A(U,V) = \{\Theta  \in \mathbb{R}^{d_1 \times d_2} \,:\, \text{col}(\Theta) \subseteq U,\,  \text{row}(\Theta) \subseteq V\}$.
It can then be seen that any low-rank matrix $\Theta \in \mathbb{R}^{d_1 \times d_2}$ of rank $k \le \min(d_1,d_2)$ lies in a collection of the low-dimensional subspaces above.

A critical question in such high-dimensional settings is how to automatically extract and leverage such low-dimensional subspace structure. \citet{negahban2012unified} showed that so long as regularization functions $r(\cdot)$ satisfy a property known as decomposability with respect to one of the collections of subspaces, regularized empirical loss minimizers yield solutions that lie in a low-dimensional subspace within that collection. Towards defining this, they require another ingredient which is a collection of orthogonal subspaces of parameters with orthogonal structure. For sparse vectors, the orthogonal subspace $B(S) = A(S)^{\perp}$. For low-rank matrices, the orthogonal subspace is $B(U,V) = \{\Theta  \in \mathbb{R}^{d_1 \times d_2} \,:\, \text{row}(\Theta) \subseteq U^{\perp},\, \text{col}(\Theta) \subseteq V^{\perp}\}$.
It can be seen that in this case, we have that $B(U,V) \subseteq A^{\perp}(U,V)$, since we do not simply want all orthogonal parameters to the structured subspace, but want orthogonal parameters which are also structured with respect to the collection of subspaces. A regularization $r(\cdot)$ is said to be decomposable with respect to collection of subspaces if for any such structured subspace pair $(A,B)$, we have that: $r(u + v) = r(u) + r(v)\; \forall u \in A, v \in B$.
For the case of sparse vector subspaces, the $\ell_1$ norm $r(\theta) = \|\theta\|_1$, and for the case of low-rank matrices, the nuclear norm $r(\Theta) = \|\Theta\|_*$ can be shown to be decomposable~\citep{negahban2012unified}.

The sub-differential of the regularization function can be written as:
$\partial r(\theta) = \{u \,|\, r(\theta') - r(\theta) \ge \langle u, \theta' - \theta\rangle, \forall \theta' \in \Theta\}$.
In the case of structured parameters above, the sub-differential in turn has additional structure. Suppose $(A,B)$ is the subspace pair corresponding to the structured parameter $\theta$. Then, for any $g \in \partial r(\theta)$, we have that $g = u_\theta + v$, where $u_\theta \in A$ has a unique representation that depends on $\theta$, and $v \in B$.
Moreover, there exists a (non-unique) inverse transform $(\partial_{\theta} r)^+$ of the partial differential, so that $(\partial_{\theta} r)^+(g) = \theta$, for all $g \in \partial r(\theta)$, with the property that $(\partial_{\theta} r)^+$ is a positive-definite linear operator, with range within the structured subspace $A$.

\section{Representer Theorem for High-Dimensional Models}

We are interested in regularized empirical risk minimizers. Given $n$ training samples $(\bx_1, y_1), (\bx_2,y_2),\cdots, (\bx_n, y_n) $ $\in \mathcal{X} \times \real$, a loss function $\ell(\cdot, \cdot): \real \times \real \rightarrow \real$, and parameters of a linear model $\theta \in \mathbf{\Theta}$, where $\mathbf{\Theta} \subseteq \mathcal{X}$ we consider the following optimization problem: 
\begin{equation}
\label{eqn:obj_unified_reg}
\hat{\theta} = \argmin_{\theta \in \mathbf{\Theta}} 
\frac{1}{n} \sum_{i=1}^n \ell(y_i, \langle x_i, \theta \rangle) + \lambda r(\theta).
\end{equation}
In the sequel, we assume that the regularization function $r(\cdot)$ is decomposable with respect to some collection of low-dimensional structured subspace, as briefly reviewed in Section~\ref{sec:prelim}. Its role is to encourage the model parameter $\theta$ to have the appropriate low-dimensional structure, while the hyper-parameter $\lambda$ balances loss and regularization.

\begin{theorem}
\label{thm:rep_unified_reg}
(high-dim representer theorem) The minimizer $\hat{\theta}$ of Eqn.\eqref{eqn:obj_unified_reg} can be written as
\begin{equation}
\hat{\theta} = \sum_{i=1}^n \left( -\frac{1}{n \lambda} \ell'(y_i, \langle x_i, \hat{\theta} \rangle) \right) \left(  (\partial_{\hat{\theta}} r)^{+} x_i \right),
\end{equation}
where $\ell' = \partial \ell /\partial ( \langle x_i, \hat{\theta} \rangle )$ denotes the partial derivative of $\ell$ with respect to its second input variable, and $(\partial_{\hat{\theta}} r)^+$ is the (non-unique) inverse transform of the regularization sub-differential. For any given test sample $x' \in \mathcal{X}$, its prediction can be decomposed according to training samples:
\begin{equation}
\label{eqn:uni_reg_decomp}
\langle x', \hat{\theta} \rangle = \sum_{i=1}^n  \underbrace{ -\frac{1}{n\lambda}   \ell'(y_i, \langle x_i, \hat{\theta} \rangle)) }_{\text{ global importance } }
\underbrace{ \langle (\partial_{\hat{\theta}} r)^{\frac{+}{2}} x_i  , (\partial_{\hat{\theta}} r)^{\frac{+}{2}} x' \rangle }_{ \text{local importance} },
\end{equation}
where $(\partial_{\hat{\theta}} r)^{\frac{+}{2}}$ is the square-root of the sub-differential inverse transform.
\end{theorem}

Eqn.\eqref{eqn:uni_reg_decomp} provides the attribution of each training sample $x_i$ to a test sample $x'$, which can be decomposed into the global importance and local importance. The global importance is a measure of how sensitive the training sample $x_i$ is to the objective and depends on the derivative of the loss function. The local importance measures the similarity between the training sample $x_i$ and the test sample $x'$. 

The local importance similarity focuses on the projection of the data points onto a structured low-dimensional subspace $A$ since the range of the sub-differential inverse transform $(\partial_{\hat{\theta}} r)^{\frac{+}{2}}$ is the structured subspace within which the parameter lies. We can thus think of such high-dimensional model estimation as specifying the local kernel $k(x,x') = \langle (\partial_{\hat{\theta}} r)^{\frac{+}{2}} x, (\partial_{\hat{\theta}} r)^{\frac{+}{2}} x' \rangle$. 
To see a crucial difference with $\ell_2$ regularized models~\citep{yeh2018representer}, where the local importance is simply an inner product between $x_i$ and $x’$, high-dimensional representers ignore the features in the orthogonal space $B$ since they have no impact on test predictions.

The theorem is derived from solving first-order optimality condition on the low-dimensional subspace $A$, i.e. one subgradient of the minimizer with respect to the objective equals zero. Next, we utilize the fact that the sub-differential $\partial r(\hat{\theta})$ has a unique representation in the model subspace $A$. It allows us to develop the inverse transform operator $(\partial_{\hat{\theta}} r)^{+}$ and use it to recover the model parameter. 

In cases where the inverse transform is non-unique, we would obtain multiple local importance, one for each inverse transform, and we can then take an average of these when computing the local importance. 

While the above development was quite abstract, in the following sections, we derive its consequences for the important settings of sparse vectors with $\ell_1$ regularization, and low-rank matrices with nuclear norm regularization.

\subsection{$\ell_1$-regularized Linear Optimization}
\label{sec:l1_representer_thm}
Based on the general theorem, we derive the representer point selection method for $\ell_1$ regularization. 
We consider the following special case of Eqn.\eqref{eqn:obj_unified_reg}:   
\begin{equation}
\label{eqn:obj_l1_reg}
\hat{\theta} = \argmin_{\theta \in \real^{p}} 
\frac{1}{n} \sum_{i=1}^n \ell(y_i, \langle x_i, \theta \rangle) + \lambda \norm{\theta}{1}, 
\end{equation}
where the $\ell_1$ regularization encourages the model  to be sparse.
Some examples of Eqn.\eqref{eqn:obj_l1_reg} include $\ell_1$-regularized generalized linear models~\citep{tibshirani1996regression}, compressed sensing~\citep{donoho2006compressed}, and sparse estimation of Gaussian graphical models~\citep{yuan2007model,friedman2008sparse}. 

We develop the representer theorem for $\ell_1$ regularized problems using Theorem \ref{thm:rep_unified_reg}. In this case, the structural model subspace is specified by the sparse model parameter $\hat{\theta}$, $A(S(\hat{\theta})) = \{ \theta \in \real^p:\theta_j = 0, \forall j \notin S \} $, where $S(\hat{\theta})$ denotes a set of coordinates that $\hat{\theta}$ has non-zero values. The orthogonal subspace $B(S(\hat{\theta}))$ is in turn a set of vectors in $\real^p$ whose coordinates on $S(\hat{\theta})$ are zero. 

Next, the sub-differential of the $\ell_1$ norm is $\partial \norm{\hat{\theta}}{1} = \{ g \in \real^p | g_i = \mbox{sign}(\hat{\theta}) \text{ if } \hat{\theta}_i \neq 0, \text{ and } |g_i| \leq 1  \text{ if } \hat{\theta}_i = 0\}$, which has a unique representation, $\mbox{sign}(\hat{\theta})$, in $A(S(\hat{\theta}))$. Next, the inverse transform can be developed by reconstructing $\hat{\theta}$ in the model subspace and zeroing out the sub-differential in the orthogonal space, where the model parameters are zero. Specifically, we use  
$(\partial_{\theta} r)^+(x) = |\hat{\theta}| \odot x, \forall x \in \real^p$, where $|\cdot|$ denotes a coordinate-wise absolute value operator, and $\odot$ denotes element-wise multiplication. Clearly, we have $(\partial_{\theta} r)^+(g) = \hat{\theta}$ for all $g \in \partial \norm{\hat{\theta}}{1}$  since $\hat{\theta}_j g_j = \hat{\theta}_j$ if $j \in S(\hat{\theta})$ and $\hat{\theta}_j g_j = 0$ if $j \notin S(\hat{\theta})$. By plugging these notations to Theorem \ref{thm:rep_unified_reg}, we obtain the following representer theorem for $\ell_1$ regularized linear optimization problems.

\begin{corollary}
\label{thm:rep_l1_reg}
(high-dim representer theorem for $\ell_1$-regularizaion) The minimizer $\hat{\theta}$ of Eqn.\eqref{eqn:obj_l1_reg} can be written as
\begin{equation}
\hat{\theta} = \sum_{i=1}^n \left( -\frac{1}{n \lambda} \ell'(y_i, \langle x_i, \hat{\theta} \rangle) \right) \left( |\hat{\theta}| \odot x_i \right),
\end{equation}
For any given test sample $x' \in \real^p$, its prediction can be decomposed according to training samples:
\begin{align}
\label{eqn:l1_reg_decomp}
\langle x', \hat{\theta} \rangle = \sum_{i=1}^n  \underbrace{ -\frac{1}{n\lambda}   \ell'(y_i, \langle x_i, \hat{\theta} \rangle)) }_{\text{ global importance } \alpha_i}
\underbrace{ \langle  \sqrt{|\hat{\theta}|} \odot x_i  , \sqrt{|\hat{\theta}|} \odot x' \rangle }_{ \text{local importance} },
\end{align}
where $\sqrt{\cdot}$ is a coordinate-wise square root operation.
\end{corollary}

With Corollary~\ref{thm:rep_l1_reg}, we can quantify training data influence on a specific test sample $(x', y')$. The sign of $\alpha_i \langle \sqrt{|\hat{\theta}|} \odot x_i, \sqrt{|\hat{\theta}|} \odot x' \rangle$ indicates whether a training sample $(x_i, y_i)$ has positive or negative influence on the test sample. Also, if a training sample $(x_i, y_i)$ has a large importance value to a test sample $x'$, two conditions must be satisfied: (1) global importance $\alpha_i$ is large (2) $\sqrt{|\hat{\theta}|} \odot x_i$ is close to $\sqrt{|\hat{\theta}|} \odot x'$. That is, $x_i$ and $x'$ are close on the coordinates where the model parameters $\hat{\theta}$ have non-zero values. 

\subsection{Nuclear-norm Regularized Linear Optimization}
\label{sec:nuclear}

We consider the following canonical nuclear norm regularized  linear optimization problem with inputs and model parameters being matrices. 
Given $n$ training samples $(X_1, y_1), \cdots, (X_n, y_n)\in \real^{d_1 \times d_2} \times \real$,
a loss function $\ell(\cdot, \cdot): \real \times \real \rightarrow \real$, and parameters of a linear model $\Theta \in \real^{d_1 \times d_2}$, we consider the following  problem: 
\begin{equation}
\label{eqn:obj_nuclear_norm_reg}
\hat{\Theta} = \argmin_{\Theta \in \real^{d_1 \times d_2}} 
\frac{1}{n} \sum_{i=1}^n \ell(y_i, \langle X_i, \Theta \rangle_F) + \lambda \norm{\Theta}{*},
\end{equation}
where $\langle \cdot, \cdot \rangle_F$ is a Frobenius inner product operator, and $\norm{\cdot}{*}$ is the 
Nuclear norm, defined as the sum of $\ell_1$ norm of singular values.  
This formulation has been applied in matrix completion~\citep{candes2010power}, matrix regression~\citep{yang2016nuclear}, and matrix compressed sensing~\citep{eldar2012compressed} with low-rank constraints. 

As in \citet{negahban2012unified}, the low-rank model subspace $A(U, V)$ is specified by a full singular value decomposition (SVD) of the model parameter $\hat{\Theta} = U\Sigma V^{\top}$, where the columns of $U \in \real^{d_1 \times k}$ and $V \in \real^{d_2 \times k}$ are orthogonal, $\Sigma \in \real^{k \times k}$ is a diagonal matrix, and $k = \mbox{rank}(\hat{\Theta})$. The orthogonal subspace is $B(U,V) = \{\Theta  \in \mathbb{R}^{d_1 \times d_2} \,:\, \text{row}(\Theta) \subseteq U^{\perp},\, \text{col}(\Theta) \subseteq V^{\perp}\}$. 

The sub-differential of the nuclear norm~\citep{watson1992characterization} is
$\partial \norm{\hat{\Theta}}{*} = \{ 
UV^{\top} + W : W \in \real^{d_1 \times d_2}, \norm{W}{2} \leq 1, W V = \mathbf{0}, U^{\top}W = \mathbf{0}
\}$, which can be decomposed as a unique representation in the model subspace ($UV^{\top} \in A(U, V)$) and $W \in B(U, V)$ in the orthogonal space. In this case, the inverse transform of sub-differential is not unique: it can be either $(\partial_{\hat{\theta}} r)^+(X) = U\Sigma U^{\top}X$ or $(\partial_{\hat{\theta}} r)^+(X) =X V\Sigma V^{\top}$ for any $X \in \real^{d_1 \times d_2}$. One can easily verify that the inverse transform recovers $\hat{\Theta}$, $(\partial_{\hat{\theta}} r)^+(\partial \norm{\hat{\Theta}}{*} ) = \hat{\Theta}$ using the fact that $U^{\top}U = VV^{\top} = I_k$. By instantiating the inverse transform to Theorem \ref{thm:rep_unified_reg}, we obtain the following corollary.
 
\begin{corollary}
\label{thm:rep_nuc_reg}
(high-dim representer theorem for nuclear-norm regularizaion) Let $U\Sigma V^{\top} = \hat{\Theta}$ be a full SVD of the minimizer $\hat{\Theta}$ of Eqn.\eqref{eqn:obj_nuclear_norm_reg}.
The minimizer of Eqn.\eqref{eqn:obj_nuclear_norm_reg} can be written as 
\begin{align}
\hat{\Theta}
= \sum_{i=1}^n - \frac{1}{n\lambda}   \ell'(y_i, \langle X_i, \hat{\Theta} \rangle_F)
\left(  U\Sigma U^{\top} X_i \right) = \sum_{i=1}^n - \frac{1}{n\lambda}   \ell'(y_i, \langle X_i, \hat{\Theta} \rangle_F)
\left(   X_i V^{\top}\Sigma V \right).
\end{align}
For any given test sample $X' \in \real^{d_1 \times d_2}$, its prediction can be decomposed according to training samples:
\begin{align}
\label{eqn:nuc_reg_decomp1}
\langle X', \hat{\Theta} \rangle 
& \!=\! \sum_{i=1}^n  - \frac{1}{n\lambda}   \ell'(y_i, \langle X_i, \hat{\Theta} \rangle))  
\langle  \sqrt{\Sigma} U^{\top} X_i  , \sqrt{\Sigma} U^{\top} X' \rangle_F \\ 
\label{eqn:nuc_reg_decomp2}
& \!=\! \sum_{i=1}^n  - \frac{1}{n\lambda}   \ell'(y_i, \langle X_i, \hat{\Theta} \rangle))  
\langle   X_i V \sqrt{\Sigma} , X' V \sqrt{\Sigma}  \rangle_F,
\end{align} 
where $\sqrt{\Sigma} = \mbox{diag}[\sqrt{\Sigma_{11}}, \cdots, \sqrt{\Sigma_{kk}}]$.
\end{corollary}
Again, the first term in Eqn.\eqref{eqn:nuc_reg_decomp1} and Eqn.\eqref{eqn:nuc_reg_decomp2}, $ - \frac{1}{n\lambda}   \ell'(y_i, \langle X_i, \hat{\Theta} \rangle))$, is the \emph{global importance} and the second inner product terms
are the \emph{local importance}. We first project input matrices $X_i$ and $X'$ onto the column or row spaces by multiplying with  $\sqrt{\Sigma}U^{\top}$ or $V\sqrt{\Sigma}$, respectively,  and then computing the Frobenius inner product. 
This term measures local similarities between a test sample and training samples in the column or row spaces of the minimizer $\hat{\Theta}$. 

Unlike Corollay \ref{thm:rep_l1_reg},
Eqn.\eqref{eqn:nuc_reg_decomp1} and Eqn.\eqref{eqn:nuc_reg_decomp2} provide two distinct ways to decompose the learned model, leading to two different ways for data attribution.
We refer to Eqn.\eqref{eqn:nuc_reg_decomp1} as \emph{column-based attribution} and Eqn.\eqref{eqn:nuc_reg_decomp2} as \emph{row-based attribution}, since they compute local importance on the column/row spaces of $\hat{\Theta}$, respectively. The interpretation of these two attributions may depend on applications. For example, as we will show in Corollary~\ref{cor:rep_mf}, the two attributions correspond to user-based attribution and item-based attributions when $U$ and $V$ are user and item embeddings in recommender systems. In other cases, we may take the average of the two local importance.

\subsection{Computation of High-dimensional Representers}
In this section, we introduce the computation of high-dimensional representers. To explain a model's prediction on $x'$, one needs to compute the high-dimensional representers for the test sample $x'$ with respect to all training samples $\{(x_i,y_i)\}_{i=1}^n$. In practice, we could pre-process the training to accelerate the computation. Recall that high-dimensional representers in Eqn.\eqref{eqn:uni_reg_decomp} consist of two components: a global importance $ \alpha_i = -\frac{1}{n\lambda}   \ell'(y_i, \langle x_i, \hat{\theta} \rangle))$, and a local importance $\langle (\partial_{\hat{\theta}} r)^{\frac{+}{2}} x_i  , (\partial_{\hat{\theta}} r)^{\frac{+}{2}} x' \rangle$. At the pre-processing step, we compute the global importances $\alpha$ for all training samples and their projections onto the low-dimensional model space, i.e. $(\partial_{\hat{\theta}} r)^{\frac{+}{2}} x_i$ for all $i \in [n]$. 

Note that global importances can be obtained by inferring all training data and calculating their derivatives. The projection operator can usually be obtained from the training stage since the model parameter $\hat{\theta}$ is available in the $\ell_1$ case, and the full SVD can usually be obtained from the training stage~\citep{mazumder2010spectral} in the nuclear norm case. The pre-processing step requires $O(np)$ and $O(nkd_1d_2)$ time for the $\ell_1$-norm and nuclear norm cases respectively. We note that in the nuclear-norm case, the pre-processing step typically takes no longer than training the regularized models with a single epoch. This is because the training samples typically need to be projected to the low-dimensional space to calculate the update formula~\citep{hsieh2014nuclear}.

Next, to explain a test prediction, we need to (1) project the test sample to the model subspace and (2) compute the inner product between the test and training samples in the model subspace. While step (1) only needs to tackle one sample, step (2) takes $O(np)$ and $O(n\max(d_1,d_2)k)$ time for the $\ell_1$-norm and nuclear norm cases respectively.

In many applications of sample-based explanations, such as generating human-understandable explanations, we only care about the top influential samples for a test prediction. This can be significantly sped up by approximate nearest neighbor search algorithms which can be run in sublinear time since we only need to find training samples with the highest inner product values.

\section{Applications to Collaborative Filtering (CF)}
\label{sec:cf}

With the widespread deployment of recommender systems across various online platforms, the significance of explainable recommender systems has grown substantially~\citep{zhang2020explainable}. Studies have indicated that users prefer recommendations that are explainable, and explanation tools are vital for debugging recommendation models~\citep{tintarev2007survey}. In this section, we showcase how high-dimensional representers can effectively explain collaborative filtering models and (deep) recommender systems.

\paragraph{Notations:}
Given a set of users $\calU$, a set of items $\calI$ and a set of user-item interactions $\calD = \{ (i,j) | \mid  i \in \calU, j \in \calI, y_{ij} \text{ is observed }\}$, CF aims to learn a $k$-dimensional embedding for each user and item, and utilizes inner products between user and item embeddings to predict unknown elements in the matrix.

\subsection{Matrix Factorization (MF) with Nuclear Norm Regularization}
Matrix factorization with nuclear norm regularizations~\citep{candes2010power, candes2012exact} is a successful model in CF. Given an incomplete rating matrix $Y \in \real^{|\calU| \times |\calI|}$ with each entry $Y_{ij}= y_{ij}, \ \forall \  (i,j) \in \calD$, the model assumes that the rating matrix $Y$ is low-rank and solves the following optimization problem:
\begin{equation}
\label{eqn:mf_with_nuc_reg}
\hat{\Theta} = \argmin_{\Theta \in \real^{|\calU| \times |\calI|}} \frac{1}{|\calD|} \sum_{(i,j) \in \calD} \ell(y_{ij}, \Theta_{ij}) + \lambda \norm{\Theta}{*},
\end{equation}
where $\hat{\Theta} \in \real^{|\calU| \times |\calI|}$ is a predicted low-rank rating matrix, $\ell(\cdot, \cdot)$ is a loss function such as square loss, and
$\lambda$ is the regularization parameter.

We apply Corollary \ref{thm:rep_nuc_reg} to Eq.\eqref{eqn:mf_with_nuc_reg} to obtain sample-based explanations. 
We represent each training pair $(i,j)$ by a matrix $X \in \real^{|\calU| \times |\calV|}$ which contains only one
nonzero entry $X_{ij} =1$, so that $\langle X, \Theta \rangle_F = \Theta_{ij}$. The resulting theorem is as below:

\begin{corollary}
\label{cor:rep_mf}
(high-dim representers for matrix factorization) Let $\hat{\Theta}$ be the minimizer of Eqn.\eqref{eqn:mf_with_nuc_reg} with $\mbox{rank}(\hat{\Theta}) = k$. Let $U \Sigma V^{\top} = \hat{\Theta}$ be its full SVD decomposition. For any test sample $(i',j')$ with $1 \leq i' \leq |\calU|$ and $1 \leq j' \leq |\calI|$, its prediction can be decomposed according to training samples:
\begin{align}
\label{eqn:mf_pred_decomp1}
\hat{\Theta}_{i'j'}
& = \sum_{i:(i,j') \in \calD} - \frac{1}{\lambda|\calD|} \ell'( y_{ij}, \hat{\Theta}_{ij}) \langle \sqrt{\Sigma}U_i, \sqrt{\Sigma}U_{i'}\rangle \\
\label{eqn:mf_pred_decomp2}
& = \sum_{j:(i',j)\in \calD} - \frac{1}{\lambda|\calD|} \ell'( y_{ij}, \hat{\Theta}_{ij})\langle \sqrt{\Sigma}V_j, \sqrt{\Sigma}V_{j'} \rangle,
\end{align} 
where $\sqrt{\Sigma} = \mbox{diag}[\sqrt{\Sigma_{11}}, \cdots, \sqrt{\Sigma_{kk}}]$, $U_i\in \real^{k \times 1}$ and $V_j \in \real^{k \times 1}$ denote $i^{th}$ and $j^{th}$ row of $U$ and $V$ respectively.
\end{corollary}
Corollary~\ref{cor:rep_mf} shows that the predicted score between user $i'$ and item $j'$, $\hat{\Theta}_{i'j'}$, can be represented as the sum of attributions to each observed interaction $(i,j)\in \calD$. Specifically, 
Eqn.\eqref{eqn:mf_pred_decomp1} decomposes predictions according to other users interacted with the same item $j'$, while Eqn.\eqref{eqn:mf_pred_decomp2} decomposes predictions according to other items interacted with the same user $i'$, 
They are referred to as  \emph{user-based attributions} and \emph{item-based attributions}, respectively. Also, we can observe that a test sample $(i',j')$ is only relevant to training samples with the same user $i'$ or the same item $j'$. 
Combining the two attributions, we define the importance score of each training data to a test sample $(i', j')$ as follows:

\begin{definition}
\label{def:L1_representers}
(high-dim representers for CF)
The importance of a training point $(i, j) \in \calD$ to a test sample $(i', j')$, $\mathbf{I}((i,j),(i',j'))$,  is given by 
\begin{equation}
\begin{cases}
-\frac{1}{\lambda|\calD|}\ell' ( y_{ij}, \langle \tilde{U}_i, \tilde{V}_{j} \rangle ) \ \langle \tilde{U}_i, \tilde{U}_{i'} \rangle
& \text{if j = j'.} \\
-\frac{1}{\lambda|\calD|}\ell'( y_{ij}, \langle \tilde{U}_i, \tilde{V}_{j} \rangle  )\ \langle \tilde{V}_j, \tilde{V}_{j'} \rangle  & \text{if i = i'.}  \\
0 & \text{ otherwise.}\\
\end{cases}
\end{equation}
where $\tilde{U} = U\sqrt{\Sigma}$ and  $\tilde{V} = V\sqrt{\Sigma}$ are normalized embedding matrices for user and item respectively. 
\end{definition}

Note that we replace $\hat{\Theta}_{ij}$ with $\langle \tilde{U}_i, \tilde{V}_{j} \rangle$ as they are equivalent. 
If a training sample $(i, j)$ has a large importance score, three conditions must be satisfied: (1) It has the same user or item as the test sample. (2) $|\ell' ( y_{ij}, \langle \tilde{U}_i, \tilde{V}_{j} \rangle )|$ must be large. When the loss function $\ell(\cdot,\cdot)$ is strongly convex, it implies that the training sample incurs a large loss. (3) Their normalized user (or item) embeddings are close. 

\subsection{General Matrix-factorization-based Models}
\label{sec:gen_mf_model}
Instead of using the nuclear norm, many matrix factorization methods directly reparameterizing the 
rating matrix $\Theta$ with the product of two low-rank matrices $U$ and $V$~\citep{koren2009matrix, mao2021simplex}, corresponding to user and item embeddings. They then directly solve the following 
optimization problem:
\begin{equation}
\label{eqn:general_mf}
\hat{U}, \hat{V} = \argmin_{U \in \real^{|\calU| \times k }, V \in \real^{|\calI| \times k }}
\sum_{(i,j) \in \calD} \ell(y_{ij}, \langle U_i, V_j \rangle),
\end{equation}
where the loss function $\ell(\cdot,\cdot)$ is point-wise, and training data $\calD$ may include negative samples for implicit CF. Popular choices include binary cross-entropy (BCE)~\citep{he2017neural}, mean square error (MSE)~\citep{fang2020influence}, and triplet loss~\citep{dahiya2021siamesexml}.  

Theorem~\ref{thm:rep_nuc_reg} does not apply to this formulation since it does not have nuclear norm regularization. While it is possible to replace $UV^{\top}$ with $\Theta$ and retrain the model with nuclear norm regularization, the retrained model may behave differently compared to the given model. 
However, the formulation does enforce hard low-rank constraints on the rating matrix through reparameterization. Therefore, to conduct sample-based attribution, we assume  Eqn.\eqref{eqn:general_mf} is implicitly regularized and use Definition~\ref{def:L1_representers} to obtain the high-dimensional representer. For this formulation, we drop the constant term, $1/\lambda|\calD|$, since $\lambda$ is unavailable and does not affect relative importance among training samples. The process of computing the high-dimensional representer for CF and its time complexity analysis are provided in Section~\ref{sec:computation} in the supplementary material.

\subsection{Two-tower models}
Two-tower networks are widely used in deep recommender systems~\citep{hidasi2015session,covington2016deep, mao2021simplex, li2019multi}. 
They encode user information and item information with two separate neural networks, which are called towers. 
The user tower maps each user (e.g., user history, features, and id) to 
a $k$-dimensional user embedding, while the 
item tower maps each item (e.g., product description and id) to 
the same embedding space. The prediction score is then calculated by the inner product of the user and item embeddings.
Formally, let the two separate towers be $f_{\theta_1}$ and $g_{\theta_2}$. The  training objective function can be written as:
\begin{equation}
\label{eqn:two_tower_obj}
\hat{\theta}_1, \hat{\theta}_2 = \argmin_{\theta_1, \theta_2} \sum_{(i,j) \in \calD} \ell(y_{ij}, \langle f_{\theta_1}(u_i), g_{\theta_2}(v_j) \rangle),
\end{equation}
where $u_i$ and $v_j$ denote features of user $i$ and item $j$. 
Again, we focus on models trained with point-wise loss functions.

To explain two-tower models, we consider the final interaction layers as a bilinear matrix factorization model and the remaining layers as fixed feature encoders. Then we apply the same explanation technique as MF models to explain them. Specifically, we concatenate embeddings of all users and items to form a user matrix and an item matrix, i.e.
\begin{align}
\nonumber
\hat{U} & = [f_{\hat{\theta}_1}(u_1); \cdots ; f_{\hat{\theta}_1}(u_{|\calU|})] \in \real^{|\calU| \times k} \\
\label{eqn:two_tower_matrices} 
\text{ and } \hat{V} & = [g_{\hat{\theta}_2}(v_1);\cdots; g_{\hat{\theta}_2}(v_{|\calI|})] \in \real^{|\calI| \times k}.
\end{align}
Then we use Definition~\ref{def:L1_representers} to obtain its sample-based explanations. 
\section{Experimental Results}
\label{sec:exp}
We perform experiments on multiple datasets to validate that 
the proposed method is a preferable choice compared with other sample-based explanation methods such as $\ell_2$ representer point selection and influence function, under the high dimensional setting. 
Moreover, we showcase the utility of the high-dimensional representer in understanding predictions of recommender systems. We also provide another use case for improving negative sampling strategies for collaborative filtering in Appendix~\ref{sec:exp_neg_sampling} and additional comparisons with other approaches in Appendix~\ref{sec:compare_tracin}.

\subsection{Evaluation Metrics} 
For quantitative evaluations, we use \emph{case deletion diagnostics}~\citep{yeh2022first,han2020explaining,cook1982residuals} as our primary evaluation metric. 
This metric measures the difference in models' prediction score at a particular test sample $z'$ after removing (a group of) influential training samples and retraining whole models. This metric helps validate the efficacy of sample-based explanation methods and provides a quantitative measurement.

We denote two metrics as $\text{DEL}_{+}(z', k, \mathbf{I})$ and $\text{DEL}_{-}(z', k, \mathbf{I})$ separately. These two metrics measure \emph{the difference between models' prediction scores when we remove top-k positive (negative) impact samples given by method $\mathbf{I}$ and the prediction scores of the original models.}
We expect $\text{DEL}_{+}$ to be negative and $\text{DEL}_{-}$ to be positive since models' prediction scores should decrease (increase) when we remove positive (negative) impact samples.

To evaluate deletion metric at different $k$, we follow~\citet{yeh2022first} and report area under the curve (AUC):
\begin{align*}
\text{AUC-DEL}_{+} = \sum_{i=1}^m \frac{\text{DEL}_{+}(z', k_i, \mathbf{I})}{m}, \text{AUC-DEL}_{-} = \sum_{i=1}^m \frac{\text{DEL}_{-}(z', k_i, \mathbf{I})}{m},
\end{align*}
where $k_1 < k_2 <\cdots < k_m$ is a predefined sequence of $k$.

\subsection{Quantitative Evaluation on $\ell_1$-regularized Models}
\label{sec:exp_glm}
In this section, we evaluate the effectiveness of the high-dimensional representer in explaining $\ell_1$-regularized logistic regression.

\subsubsection{Experimental Settings}
\paragraph{Datasets and models being explained:} We use the following three datasets on binary classification. \textbf{(1) 20 newsgroups\footnote{http://qwone.com/~jason/20Newsgroups/}:} This dataset contains roughly $20,000$ newsgroups posts on 20 topics.  It contains $19,996$ samples with $1,355,191$ features. We randomly split $10\%$ data for the test set. \textbf{(2) Gisette~\citep{guyon2004result}:} It is a handwritten digit recognition problem, which contains highly confusible digits '4' and '9'. It contains $6,000$/$1,000$ samples with each containing $5,000$ features for training/testing. \textbf{(3) Rcv1~\citep{lewis2004rcv1}:} It is a benchmark dataset on text categorization. It has $20,242$/$677,399$ samples for training/testing. We use bag-of-words features with dimensions $47,236$. We train logistic regression models with $\ell_1$ regularization using LIBLINEAR~\citep{fan2008liblinear} on the three datasets. The accuracy of models on the three datasets is above $97\%$.

\paragraph{Baselines:}
We compare the high-dimensional representer with the $\ell_2$ representer, the influence function (IF) and random deletions. Given a test sample $x'$, the $\ell_2$ representer calculates importance score of a training point $(x_i,y_i)$ to the test sample $x'$ with the following formula:
$$
\mathbf{I}_{\ell_2}((x_i,y_i), x') = -\ell'(y_i, \langle x_i, \hat{\theta} \rangle))  \langle  x_i  , x' \rangle.
$$
For the influence function, we adopt the formula in Proposition 5.3 of \citet{avella2017influence}. Assume only the first $q \leq p$ entries of the minimizer $\hat{\theta}$ are nonzero, the influence function, $\mathbf{I}_{IF}((x_i,y_i), x')$, is given by
\begin{align*} 
-( \frac{1}{n}\nabla_{\theta_{1:q}}  \ell(y_i, \langle x_i, \hat{\theta} \rangle) + \lambda \mbox{sign}(\hat{\theta})_{1:q})^\top H_{\hat{\theta}_{1:q}}^{-1}  x'_{1:q},
\end{align*}
where $H_{\hat{\theta}_{1:q}} = \sum_{i=1}^n \nabla^2_{\theta_{1:q}}  \ell(y_i, \langle x_i, \hat{\theta} \rangle) \in \real^{q \times q}$. The calculation of the influence function can be simply viewed as first projecting features $x$, $x'$, and the model parameter $\hat{\theta}$ to nonzero entries of $\hat{\theta}$ and then computing the influence function normally. Notice that the naive implementation takes $O(nq^3 + np)$ time complexity to compute inverse hessian matrix, while the high-dim and $\ell_2$ representers only take $O(np)$ to compute importance scores of all training samples to a test prediction.

To compute $\text{AUC-DEL}$ scores, we set $ k_i = 0.01i N$ for $1 \leq i \leq 5$. We remove $1\%$ to $5\%$ of positive (negative) impact training samples and report the averaged prediction difference after removing these samples. Each metric is reported over $40$ trials with each trial containing $40$ test samples.

\subsubsection{Results}
The results of the four methods are presented in 
Table \ref{table:evaluation_l1_models}. We also report the averaged runtime of computing the importance of one test prediction to all training data on a single CPU. The results show that the high-dimensional representer outperforms the other three methods and is over 25x faster than the influence function. Also, the $\ell_2$ representer is slightly faster than the high-dimensional representer  since inner product is fast when the training data is sparse, and the high-dimensional representer  requires one extra step to project vectors to low-dimensional model subspace.

\begin{table}[h!]
\small
\center
\resizebox{0.5\linewidth}{!}{
\begin{tabular}{|c|c|c|c|} 
\toprule
Datasets &  20 newsgroups & Gisette & Rcv1 \\
\midrule
\midrule
\multicolumn{4}{|c|}{ $\text{AUC-DEL}_{+} $ }
\\
\midrule
High-dim Rep.  &  $\mathbf{-3.733}  \pm 0.093$ 
& $\mathbf{-1.000} \pm 0.081$ & $ \mathbf{-3.208} \pm 0.060$    \\
\midrule
$\ell_2$ Rep.  &  $-2.472\pm 0.067$ 
& $ -0.577 \pm 0.073$ & $ -2.780 \pm 0.057$   \\
\midrule
IF  &  $-2.583 \pm 0.043$ &
$ -0.531  \pm 0.011$ & $ -2.652 \pm 0.040$   \\
\midrule
Random  &  $0.006 \pm 0.014$ & $ 0.010 \pm 0.022$ & $ 0.009 \pm 0.005$   \\
\midrule
\midrule
\multicolumn{4}{|c|}{$\text{AUC-DEL}_{-} $ }
\\
\midrule
High-dim Rep.  &  $\mathbf{7.478}  \pm 0.194$ 
& $ \mathbf{3.116} \pm 0.110$ & $ \mathbf{3.170} \pm 0.077$   \\
\midrule
$\ell_2$ Rep.  &  $5.214  \pm 0.143$ 
& $ 2.118 \pm 0.093$ & $ 2.726 \pm 0.067$   \\
\midrule
IF  &  $4.894 \pm 0.086$ 
& $ 0.523 \pm 0.013 $  & $ 3.065 \pm 0.082$  \\
\midrule
Random  &  $0.003  \pm 0.014$ 
& $ 0.007 \pm 0.024$ & $ 0.007 \pm 0.005$   \\
\midrule
\midrule
\multicolumn{4}{|c|}{ Runtime (ms)}
\\
\midrule
High-dim Rep.  &  $ 61.35  \pm 0.59  $ & $ \mathbf{87.34} \pm 0.71  $  & $ 10.61 \pm 0.13 $  \\
\midrule
$\ell_2$ Rep.  &  $\mathbf{59.47} \pm 0.58 $ & $130.16 \pm 0.34 $ & $ \mathbf{6.14} \pm 0.22  $  \\
\midrule
IF  &  $2678.38 \pm 3.19 $ & $ 3628.70 \pm 2.007 $  & $ 263.90 \pm 1.01$  \\
\bottomrule
\end{tabular}}
\caption{Case deletion diagnostics for removing positive (negative) impact training samples on various datasets and models and run time comparison. $95\%$ confidence interval of averaged deletion diagnostics on $40 \times 40 = 1,600$ samples is reported. Averaged runtimes over $100$ samples are also reported. 
Smaller (larger) $\text{AUC-DEL}_{+}$ ($\text{AUC-DEL}_{-}$) is better. }
\label{table:evaluation_l1_models}
\end{table}

\begin{table*}[h!]
\small
\center
\begin{tabular}{|c|c|c|c|c|c|} 
\toprule
\multirow{2}*{Datasets} & \multirow{2}*{Models} & \multirow{2}*{Metrics} & 
\multicolumn{3}{c|}{Methods} \\
\cmidrule{4-6}
&&&  High-dim  Rep. & FIA & Random \\
\midrule
\multirow{6}*{\begin{tabular}{c}MovieLens-\\1M\end{tabular}} 
& \multirow{2}*{\begin{tabular}{c}MF w. nucl-\\ear norm \end{tabular}}
& $\text{AUC-DEL}_{+} $ &  $\mathbf{-0.225}  \pm 0.006$ & -
& $-0.002\pm 0.002$   \\
&& $\text{AUC-DEL}_{-} $ & $\mathbf{0.160}  \pm 0.004$ & -
& $-0.002 \pm 0.002$  \\
\cmidrule{2-6}
& \multirow{2}*{MF} 
& $\text{AUC-DEL}_{+} $ &  $\mathbf{-0.196}  \pm 0.006$ & $-0.101 \pm 0.004$
& $-0.002 \pm 0.002$   \\
&& $\text{AUC-DEL}_{-} $ & $\mathbf{0.169}  \pm 0.004$  & $0.072 \pm 0.004$
& $-0.001 \pm 0.002$  \\
\cmidrule{2-6}
& \multirow{2}*{\begin{tabular}{c}Youtube-\\Net \end{tabular}} 
& $\text{AUC-DEL}_{+} $ &  $\mathbf{-0.227}  \pm 0.008$ &  $-0.096 \pm 0.006$
& $-0.001 \pm 0.004$   \\
&& $\text{AUC-DEL}_{-} $ & $\mathbf{0.214}  \pm 0.007$ & $0.113 \pm 0.007$
& $0.006 \pm 0.004$  \\
\midrule
\multirow{4}*{\begin{tabular}{c}
Amazon \\ reviews 2018  \\ (video games)  
\end{tabular}} 
& \multirow{2}*{MF} 
& $\text{AUC-DEL}_{+} $ &  $\mathbf{-0.184}  \pm 0.012$ & $-0.123 \pm 0.011$
& $-0.070 \pm 0.011$   \\
&& $\text{AUC-DEL}_{-} $ & $\mathbf{0.080} \pm 0.012$ & $-0.009 \pm 0.012$
& $-0.077 \pm 0.011$  \\
\cmidrule{2-6}
& \multirow{2}*{\begin{tabular}{c}Youtube-\\Net \end{tabular}} 
& $\text{AUC-DEL}_{+} $ &  $\mathbf{-0.234}  \pm 0.014$ &  $-0.056 \pm 0.013$
& $-0.032 \pm 0.011$   \\
&& $\text{AUC-DEL}_{-} $ & $\mathbf{0.294}  \pm 0.011$  & $0.069 \pm 0.013$
& $-0.032 \pm 0.011$  \\
\bottomrule
\end{tabular}
\caption{Case deletion diagnostics for removing positive (negative) impact training samples on various datasets and models. $95\%$ confidence interval of averaged deletion diagnostics on $40 \times 40 = 1,600$ samples is reported.
Smaller (larger) $\text{AUC-DEL}_{+}$ ($\text{AUC-DEL}_{-}$) is better.}
\label{table:eval_CF}
\end{table*}

\subsection{Quantitative Evaluation on Collarborative Filtering  }
\label{sec:exp_cf}
In this section, we evaluate the effectiveness of the high-dimensional representer  on explaining CF models in recommender systems.

\subsubsection{Experimental Settings} 
\paragraph{Datasets:} \textbf{(1) Movielens-1M~\citep{harper2015movielens}:} It contains about 1M ratings (1-5) from 6,040 users on 3,706 movies. \textbf{(2) Amazon review (2018)~\citep{ni2019justifying}:} This dataset contains reviews and ratings (1-5) of products on Amazon. Since the whole dataset is too large, we use data in the video games category, which contains 284,867 ratings from 15,517 users to 37,077 items.
We follow the preprocessing procedure in \citet{cheng2019incorporating}.
We filter out users and items with less than 10 interactions. For every user, we randomly held out two items' ratings to construct the validation and test sets.
Also, we normalize all ratings to $[-1, 1]$. 
\vspace{-0.2cm}
\paragraph{Models being explained:} We test the high-dimensional representer on three different models: (1) Matrix factorization with nuclear norm regularization (MF w. nuclear norm) as in Eqn.\eqref{eqn:obj_nuclear_norm_reg}. We do not run this model on Amazon review dataset because the rating matrix is too large. (2) Matrix Factorization (MF) as in Eqn.\eqref{eqn:general_mf}. (3) YoutubeNet~\citep{covington2016deep}, which uses a deep neural network to encode user features and is one of the representative deep two-tower models.

All models are trained with squared loss.
We use soft-impute~\citep{mazumder2010spectral} algorithm to train the Model (1). The models (2) and (3) are optimized by stochastic gradient descent. Hyper-parameters and model structures are detailed in Appendix \ref{sec:exp_details}.

\vspace{-0.2cm}

\paragraph{Baselines:} We compare the high-dimensional representer with the following three baselines: (1) Fast influence analysis (FIA): since the influence function is not scalable to the size of common recommender system benchmarks, \citet{cheng2019incorporating} propose FIA as an approximation of the influence function for MF-based models.
(2) Random deletion, which we randomly delete training samples with the same user or item as the given test sample.

Notice that the FIA are not applicable to the MF with nuclear norm model since it is only applicable to MF models in Eqn.\eqref{eqn:general_mf}. Also, the $\ell_2$ representer is not applicable to models with two separate encoders since it cannot be treated as a linear mapping. We leave the comparison to TracIn~\citep{pruthi2020estimating}, which is only applicable to models trained with SGD-based optimizers, to the supplementary material. 

\subsubsection{Setup} 
We combine user-based and item-based explanations and sort them according to their importance scores. For MovieLens-1M, we drop $k=10,20,30,40,50$ samples. For Amazon reviews, we drop $k=3,6,9,12,15$ samples.
Each metric is averaged over $40$ trials with each trial having $40$ test samples.

\subsubsection{Results} 
Table~\ref{table:eval_CF} summarises the results of different methods. First, we observe that randomly removing samples have roughly no/negative  effects on models' predictions for MovieLens-1M/Amazon reviews, and all other methods outperform the random deletiton baseline.
Second, the high-dimensional representer outperforms FIA and random deletion in all settings, indicating that the high-dimensional representer is able to estimate the importance of each training sample more accurately.

\subsection{Use Case 1 : Explaining Recommender Systems' Predictions}
\label{sec:expl}

\begin{table}[h!]
\small
\center
\resizebox{0.5\linewidth}{!}{
\begin{tabular}{|c|c|c|c|} 
\toprule
Movies & User's rating & 
 Movie genre & Importance \\
\midrule
Men in Black & 3 &
\begin{tabular}{@{}c@{}}Action,Sci-Fi,  \\ Comedy,Adventure \end{tabular} & -4.55 \\
\midrule
Diabolique & 2 & Drama/Thriller & 
 -4.03 \\
\midrule
\begin{tabular}{@{}c@{}}Independence \\ Day (ID4) \end{tabular} 
& 5 & \begin{tabular}{@{}c@{}}Action,Sci-Fi, \\ War \end{tabular} 
 & 3.52\\
\midrule 
\begin{tabular}{@{}c@{}}Star Trek IV: \\ The Voyage Home  \end{tabular} 
& 5 & 
\begin{tabular}{@{}c@{}}Action,Sci-Fi,  \\ Adventure \end{tabular} & 3.12\\
\midrule
\begin{tabular}{@{}c@{}}Star Trek V: \\ The Final Frontier\end{tabular} & 2 & \begin{tabular}{@{}c@{}}Action,Sci-Fi,  \\ Adventure \end{tabular} & -2.86 \\
\midrule 
\begin{tabular}{@{}c@{}}Star Trek: \\ First Contact \end{tabular}
& 5 &
\begin{tabular}{@{}c@{}}Action,Sci-Fi,  \\ Adventure \end{tabular}& 2.59\\
\bottomrule
\end{tabular}}
\caption{An example of item-based explanations. In the example, a MF model predicts one user's rating for the movie "Star Trek VI: The Undiscovered Country" to be 3.89. The genres of the movie are action, sci-fi, and adventure.}
\label{table:example}
\end{table}
In this section, we show that the high-dimensional representer generates explanations based on users' historically interacted products for collaborative filtering models.

Table~\ref{table:example} shows an example of an explanation for movie recommendations. We use an MF model trained with square loss to predict users' ratings from 1 to 5 on Movielens-100k, a smaller version of Movielens-1M. 
We first choose a user with $87$ historical ratings and predicts their rating on "Star Trek VI: The Undiscovered Country", calculate similarity scores with the high-dimensional representer  on the user's past ratings, and then sort the items according the absolute importance scores.
The explanation can be interpreted as "the MF model predicts your rating on \textit{Star Trek VI: The Undiscovered Country} to be $3.89$ mostly because of your ratings on the following six movies." 

The explanation consists of  movies with similar genres and prequels of "Star Trek VI". We see that the model learns the relations of movies from the explanation since movie names and genres are not provided during training.
Also, the user's past ratings of 2 or 3 negatively impact the prediction, and ratings of 5 have positive influence. It is reasonable since the user's preference for similar movies would impact the model's predicted ratings. Notice that the high-dimensional representer  can also be used to provide user-based explanations in terms of the influence of other users' ratings on the same movie. We do not show these explanations here since user information is lacking in most publicly available datasets. More examples can be found in Appendix \ref{sec:more_qualitative}.

\newpage
\subsection{Use Case 2: Improving Negative Sampling Strategies}
\label{sec:exp_neg_sampling}
In this experiment, we show that high-dimensional representers can be used to improve negative sampling strategies that are widely used to train collaborative filtering models for implicit signals. 

\paragraph{Motivation:} Implicit CF learns from users' behavior that implicitly affects users' preferences. For example, it may learn from the clicks of users or users' watching history. In this setting, user-item interactions usually contain only positive interactions, and practitioners usually regard all other unobserved interactions as negative samples. However, these unobserved interactions may include false negatives. For instance, users may ignore items not displayed to them, not necessarily because users dislike them. Such false negatives have been demonstrated to be harmful to models~\cite {ding2020simplify}. However, identifying false negatives is challenging since it is impossible to ask users to look over all items and mark their preferences. 

\paragraph{Proposed approach:}
We propose to measure \emph{aggregated importance scores} of negative samples to identify these false negatives.
These scores quantitatively measure the extent to which negative pairs contribute to the decrease in prediction scores for observed positive interactions. Larger aggregated importance scores indicate that the negative pair reduces the model's confidence in other known positive interactions, suggesting a higher likelihood of being a false negative.

Let $\mathcal{D} = \mathcal{P} \cup \mathcal{N}$ be the training set comprising positive interactions $\mathcal{P}$ and negative samples $\mathcal{N}$ selected through a negative sampling strategy. The aggregated importance scores are defined as follows:

\begin{equation}
\label{eqn:importance_negative_samples}
\mathbf{I}_{neg}((i,j)) = \sum_{(i',j') \in \calP} \mathbf{I}((i,j), (i',j')),
\end{equation}
where $\mathbf{I}(\cdot,\cdot)$ is the importance score provided by the high-dimensional representer as in Definition~\ref{def:L1_representers}.
$\mathbf{I}_{neg}((i,j))$ can be interpreted as the sum of importance scores of a negative sample to all positive samples in the training set.

\subsubsection{Experimental Setup}
To validate the effectiveness of high-dimensional representers in improving negative sampling strategies, we first train a base model using a normal negative sampling strategy, and then retrain the model after removing identified negative samples. We use the change in the models' performance to measure the performance of the proposed method.

\paragraph{Datasets:} We use a binarized MovieLens-100k dataset, which contains $100,000$ ratings (1-5) from 943 users on 1,682 movies. We transform user ratings into binary signals by dropping user ratings less than $4$ and treating other interactions as positive samples. In accordance with \citet{toh2010accelerated, jaggi2010simple}, we randomly selected 50\% of the ratings for training and the others for the test set. 

\paragraph{Base models:}
We first train a matrix factorization model with uniformly selected negative samples. The model is trained with binary cross entropy loss function with the following formulation:
\begin{align*}
\argmin_{\substack{U \in \real^{|\calU| \times k }, \\ V \in \real^{|\calI| \times k }}} 
& - \sum_{(i,j) \in \calP} \log(\sigma(\langle U_i, V_j \rangle))  - 0.05 \sum_{(i,j) \in \calN} \log( 1- \sigma( \langle U_i, V_j \rangle)),
\end{align*}
where $\sigma(\cdot)$ denotes a sigmoid function, 
and $\calN = \calD \backslash \calP$ contains all unknown user-item interactions. We multiply  loss functions with negative samples with $0.05$ since it improves the models' performance. 
After calculating aggregated importance scores of all negative samples, we remove the top $p\%$ samples with the least scores from $\calN$ and train a new MF model with the same objective.

\paragraph{Evaluation metrics:} In order to assess the effectiveness of the proposed methods, we utilize the following two evaluation metrics:
\begin{enumerate}
    \item Number of false negatives identified: Given the impracticality of labeling all user-item interactions, we consider only the positive interactions in the test set as potential false negatives. This metric evaluates the number of false negatives correctly identified by each method.
    
    \item Performance improvement of the base model after retraining: We measure the change in performance of the base model after removing the top $p\%$ of negative samples identified by each method. The models' performance is evaluated using the recall@20 metric on the test set~\citep{he2020lightgcn}.
\end{enumerate}

These evaluation metrics enable us to assess the ability of the proposed methods to accurately identify false negatives and quantify the improvement achieved in the performance of the base model through retraining.

\paragraph{Baselines:}
We compare the high-dimensional representer with (1) fast influence analysis, (2) loss functions, and (3) random selections. For FIA, we use importance scores provided by FIA to compute aggregated importance scores in Eqn.\eqref{eqn:importance_negative_samples}. For loss functions, we filter out top $p\%$ negative samples with highest loss. For random selection, we randomly remove $p\%$ of negative samples from $\calN$.

\subsubsection{Experimental Results}

\begin{figure}
\centering
\subfigure[\label{fig:rq4_1}]{\includegraphics[width=0.33\textwidth]{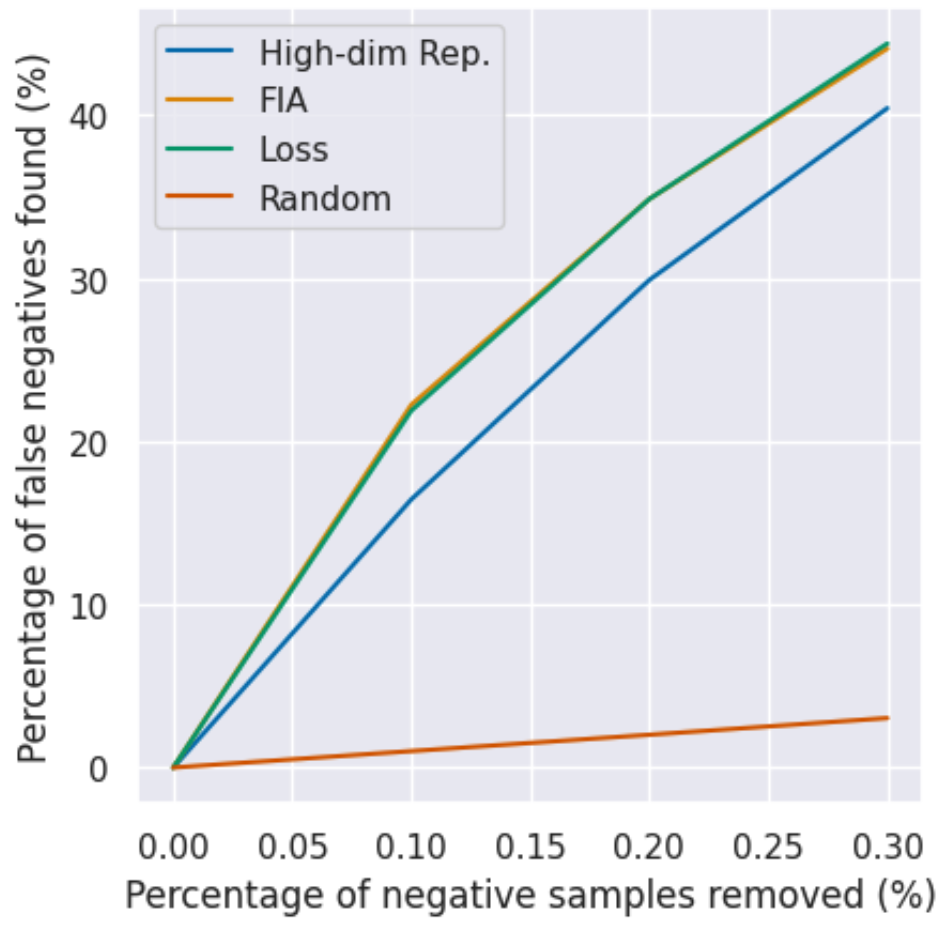} } \subfigure[\label{fig:rq4_2}]{\includegraphics[width=0.33\textwidth]{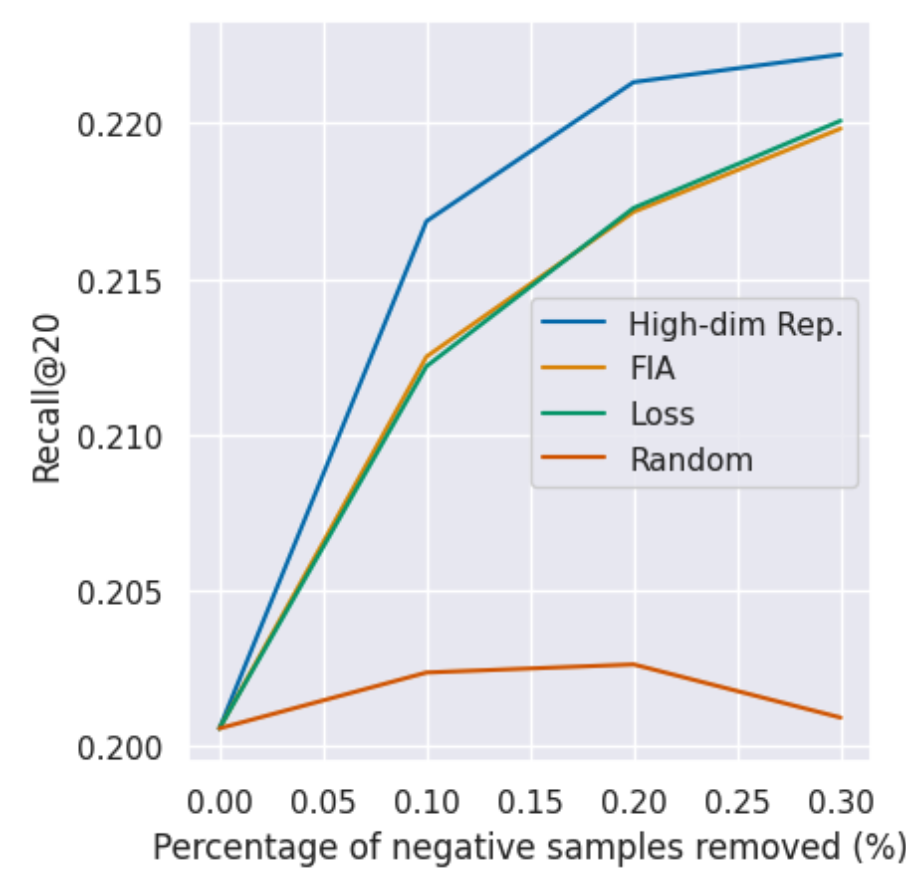}}
\caption{(a) Averaged percentage of false negative samples found. (b) Averaged models' performance improvement after removing top negative samples.}
\end{figure}

The results of our experiments are presented in Figure~\ref{fig:rq4_1} and Figure~\ref{fig:rq4_2}. We observe that the high-dimensional representer, loss functions, and FIA outperform random selection on both evaluation metrics. Notably, while the high-dimensional representer identifies slightly fewer false negatives compared to the loss functions and FIA, it identifies more influential false negatives that contribute the most to performance improvement. These findings indicate that the performance of implicit collaborative filtering can be enhanced by removing harmful samples. As a potential future direction, it would be interesting to explore the integration of the high-dimensional representer into negative sampling procedures.

\section{Conclusion}
In this paper, we present \emph{high-dimensional representers} to explain predictions of high-dimensional models in terms of contributions from each of the training samples. We investigate its consequences for canonical instances of sparse models, as well as low-rank models, together with a case study on collaborative filtering, which we consider low-rank matrix-factorization-based models as well as their deep neural variants. In future work, it would be of interest to derive corollaries of our general result for additional instances of high-dimensional models such as group-structured models, as well as additional applications such compressed sensing and sparse Gaussian graphical model estimation.

\nocite{langley00}

\bibliography{ref}
\bibliographystyle{icml2023}

\newpage
\appendix
\section{Overview of the Appendix}
The appendix is structured as follows: in Appendix \ref{sec:socail_impact}, we explore the potential social impacts of our work.
Furthermore, in Appendix \ref{sec:more_qualitative}, we present additional qualitative examples that demonstrate the use of high-dimensional representers in explaining recommender systems. Detailed experimental information regarding the experiments in Section \ref{sec:exp} can be found in Appendix~\ref{sec:exp_details}. We also provide the pseudocode and time complexity analysis of the high-dimensinoal representer for collaborative filtering in Appendix~\ref{sec:computation}. Moreover, a comparison between the high-dimensional representers for collaborative filtering and TracIn~\citep{pruthi2020estimating} is presented in Appendix~\ref{sec:compare_tracin}. Lastly, in Appendix~\ref{sec:proof}, we include proofs of our theoretical results.

\section{Potential Social Impact of Our Work}
\label{sec:socail_impact}

One potential social impact is that one may use our approach to change models' predictions via adjusting training samples. This may have positive impacts, such as debugging models or making models fair, and negative impacts, such as attacking existing models or making models more biased and unethical.

\newpage
\section{More Qualitative Examples}
\label{sec:more_qualitative}
Table \ref{table:example2} and \ref{table:example3} show qualitative examples of explaining recommender systems' predictions with high-dimensional representers. We use a matrix factorization model trained with square loss to predict users' ratings from 1 to 5 on Movielens-100k datasets.

\begin{table}[h!]
\center
\begin{tabular}{|c|c|c|c|} 
\toprule
Movies & User's rating & 
 Movie genre & Importance \\
\midrule
Henry V & 1 &
drama, war   & -13.57 \\
\midrule
Cop Land & 1 & crime, drama, mystery & 
 -11.14 \\
\midrule
Soul Food
& 5 &  drama
 & -8.67\\
\midrule 
Independence Day (ID4)
& 2 & 
action, sci-fi, war & -7.05\\
\midrule
Things to Do in Denver when You're Dead
& 5 &
crime, drama, romance & 6.86\\
\midrule
Star Trek IV:The Voyage Home  
& 2 & 
action, sci-fi, adventure & -6.66\\
\bottomrule
\end{tabular}
\caption{An example of item-based explanations. In the example, a MF model predicts one user's rating for the movie "Star Wars" to be 4.00. The genres of the movie are action, sci-fi, romance, war, and adventure.}
\label{table:example2}
\end{table}

\begin{table}[h!]
\center
\begin{tabular}{|c|c|c|c|} 
\toprule
Movies & User's rating & 
 Movie genre & Importance \\
\midrule
Terminator & 1 &
action, sci-fi, thriller   & -13.48 \\
\midrule
M*A*S*H & 1 & comedy, war & 
 -12.81 \\
\midrule
Fantasia &  1 &  animation, children, musical
 & -11.25\\
\midrule 
Psycho
& 2 & horror, romance, thriller & -10.33\\
\midrule
Batman
& 1 & action, adventure, crime, drama & -9.31\\
\midrule
Gone with the wind 
& 2 & drama, romance, war & -7.70\\
\bottomrule
\end{tabular}
\caption{An example of item-based explanations. In the example, a MF model predicts one user's rating for the movie "Top gun" to be 3.37. The genres of the movie are action and romance.}
\label{table:example3}
\end{table}

\newpage
\section{More Experimental Details}
\label{sec:exp_details}

In this section, we provide details of our experiment settings.

\subsection{$\ell_1$ regularized binary classifiers}
We detail data preprocessing for the three datasets we use in Section \ref{sec:exp_glm}, and model hyper-parameters. Dataset statistics and hyper-parameters are listed in Table \ref{table:binary_dataset_statistics}.

For the three text classification datasets in Section \ref{sec:exp_glm}, we convert these multiclass datasets into binary datasets. For 20 newsgroup, we follow the preprocessing procedure in \citet{keerthi2005modified} to group the 20 topics into two classes. For Gisette, we use the validation set as the testing set since the labels of the testing set are not available.
For Rcv1, we treat CCAT and ECAT as the positive class and treat GCAT and MCAT as the negative class. Instances in both positive and negative classes are removed.

\begin{table}[H]
\small
\centering
\begin{tabular}{|c|c|c|c|c|c|} 
\toprule
Dataset & \# of training samples  & \# of test samples & feature dimension & $n_{\text{train}}\lambda$  & Model Accuracy (\%) \\
\midrule
20 newsgroup & 17,997 & 1999 & 1,355,191 & 10 & 97.50  \\
\midrule
Gisette & 6,000 & 1,000 & 5,000 & 1 & 97.50 \\
\midrule
Rcv1  
 & 20,242 & 677,399 & 47,236 & 1 & 97.50  \\
\bottomrule
\end{tabular}
\caption{Statistics of text classification datasets, model regularization hyper-parameters, and model accuracy.  }
\label{table:binary_dataset_statistics}
\end{table}

\subsection{Collaborative Filtering}
\subsubsection{Dataset statistics}
Table \ref{table:dataset_statistics} shows the statistics of datasets we used in Section \ref{sec:exp_cf}, \ref{sec:expl} and Appendix \ref{sec:exp_neg_sampling}.

\begin{table}[H]
\small
\centering
\begin{tabular}{|c|c|c|c|c|} 
\toprule
Dataset & User \#  & Items \# & Interactions \# & Density (\%)  \\
\midrule
MovieLens-100k & 943 & 1,682 & $55,375^{\dag}$ & 3.49 \\
\midrule
MovieLens-1M & 6,040 & 3,706 & 1,000,209 & 4.47 \\
\midrule
\begin{tabular}{c}
Amazon review (2018) \\  (Video games)  \end{tabular}
 & 15,517 & 37,077 & 284,867 & 0.05  \\
\bottomrule
\end{tabular}
\caption{Statistics of different recommender system datasets. \dag We remove the ratings that are less or equal to 3, so the number of interactions becomes 55,375. }
\label{table:dataset_statistics}
\end{table}
 
\subsubsection{Implementation Details}
To evaluate the models' performance, we follow \citet{mazumder2010spectral} to use mean absolute error (MAE), which measures the absolute distance between normalized true ratings and predicted ratings. Below we detail the model architectures and their performance on different datasets. All models are trained with square error loss functions.

\paragraph{Matrix factorization with nuclear norm:} We use Soft-Impute~\citep{mazumder2010spectral} to train the models. We set max iterations to 20 and embedding dimension to 12 on the MovieLens-1M dataset. It achieves MAE of 0.54.

\paragraph{Matrix factorization:} We use SGD optimizer with learning rate 2.0/15.0 with batch size 3000/3000 to train MF model for 10/10 epochs
on MovieLens-1M/Amazon reviews 2018. The model achieves MAE of 0.36/0.46 on MovieLens-1M/Amazon reviews 2018.

\paragraph{YoutubeNet:} For MovieLens-1M/Amazon reviews 2018, we use Adam optimizer with learning rate 0.001/0.001 with batch size 3000/3000 to train YoutubeNet for 20/10 epochs. We use an embedding of 64/16 trainable parameters to model user and item information. The user feature encoder consists of 4/3 layers of size 64/16 with 0.2/0.2 dropout probabilities. The item feature encoder contains only item embeddings. 
It achieves MAE of 0.36/0.42. 

\section{Computation of the High-dimensional Representer for Collaborative Filtering}
\label{sec:computation}

The whole process of the computation of high-dimensional representers is shown in Algorithm~\ref{algo:l1_representers_for_cf}. The first step is pre-processing, which computes normalized user and item embedding matrices through computing SVD of $\hat{U}\hat{V}^{\top}$. Then we calculate $\ell_1$representers with Definition~\ref{def:L1_representers}.
We note that the shape of $\hat{U}\hat{V}^{\top}$ is $|\calU| \times |\calI|$ and it is costly to
compute SVD of it since we may have millions of users and items in real-world applications. We propose to decompose the computation into computing SVD of $\hat{U}$ and $\hat{V}$ separately and then combining these smaller matrices of size $k \times k$. We show that this decomposition significantly reduces time complexity in the following analysis.

Note that $(U_1U_3)\Sigma_3 (V_1V_3)^{\top}$ in the pre-processing step is a valid SVD of $\hat{U}\hat{V}^{\top}$ due to the fact that $(U_1U_3)^{\top}(U_1U_3) = U_3^{\top}U_1^{\top}U_1U_3 = I_{k}$ and $(V_1V_3)^{\top}V_1V_3= I_k$.

\begin{algorithm}[h]
\centering
   \caption{Computation of high-dimensional representers for Collaborative Filtering}
    \label{algo:l1_representers_for_cf}
\begin{algorithmic}
   \STATE {\bfseries Input:}  A trained user embedding matrix $\hat{U} \in \real^{|\calU| \times k}$, a trained item embedding matrix $\hat{V} \in \real^{|\calI| \times k}$, a loss function $\ell(\cdot, \cdot)$, training dataset $\calD$, and a test sample $(i',j')$.
   \STATE {\bfseries Preprocessing}{($\hat{U},\hat{V}$)}:
   \begin{ALC@g}
   \STATE $U_1,\Sigma_1', V_1^{\top} \leftarrow  SVD(\hat{U})$.
    \STATE $U_2,\Sigma_2', V_2^{\top} \leftarrow  SVD(\hat{V}^{\top})$.
    \STATE $U_3,\Sigma_3, V_3^{\top} \leftarrow  SVD(\Sigma_1V_1^{\top}U_2\Sigma_2')$. 
    \STATE $\tilde{U} \leftarrow U_1U_3 \sqrt{\Sigma_3}$, and $\tilde{V} \leftarrow V_1V_3 \sqrt{\Sigma_3}$. \\
    \STATE {\bfseries Return:} $\tilde{U},\tilde{V}$
   \end{ALC@g}
   \STATE { \bfseries Explain}$(\tilde{U},\tilde{V}, \calD, \ell, i', j')$:
    \begin{ALC@g}
    \STATE $\mbox{Expls} \leftarrow []$
    \STATE \# Computation of user-based explanations:
    \STATE $L_u \leftarrow \{(i,j') | (i,j') \in \calD  \}$ 
    \FOR{${(i,j') \in L_u}$}
        \STATE $\mbox{score} \leftarrow -\ell'(y_{ij'}, \langle \tilde{U}_{i},\tilde{V}_{j'}) \rangle \langle \tilde{U}_{i}, \tilde{U}_{i'} \rangle$.
        \STATE Append $((i,j'), \mbox{score})$ to $\mbox{Expls}$. 
    \ENDFOR
    \STATE \# Computation of item-based explanations:
    \STATE $L_v \leftarrow \{(i',j) |(i',j) \in \calD \}$
    \FOR{${ (i',j) \in L_v}$} 
        \STATE $\mbox{score} \leftarrow -\ell'(y_{i'j}, \langle \tilde{U}_{i'},\tilde{V}_{j}) \rangle \langle \tilde{V}_{j}, \tilde{V}_{j'} \rangle$.
        \STATE Append $((i',j), \mbox{score})$ to $\mbox{Expls}$. 
    \ENDFOR 
    \STATE {\bfseries Return:} $\mbox{Expls}$
    \end{ALC@g} 
\end{algorithmic}
\end{algorithm}

\subsection{Analysis of Time Complexity}
In algorithm~\ref{algo:l1_representers_for_cf}, we use a randomized singular value decomposition  algorithm~\citep{halko2011finding} to decompose matrices. The SVD algorithm takes an input matrix of size $m \times n$ and outputs its decomposition with rank $k$ using $\calO(mn\log(k) + (m+n)k^2)$ operations. 

At the pre-processing step, we perform this algorithm three times to decompose matrices of size $|\calU| \times k$, $|\calV| \times k$ and $k \times k$ to rank-$k$ matrices, which takes $O(\max(|\calU| ,|\calV| )k^2)$ time (assume that $|\calU|, |\calV| >> k$). On the contrary, if we directly perform SVD on $\hat{U}\hat{V}^{\top}$, the time complexity would be $O(|\calU||\calV|\log k)$, which is significantly larger than decomposing three smaller matrices. 
Also, the matrix multiplications on lines 4 and 5 also take $O(\max(|\calU| ,|\calV| )k^2)$ time. Therefore, the overall time complexity of the pre-processing step is $O(\max(|\calU| ,|\calV| )k^2)$.

At the explanation step, let the average number of interactions a user or an item has is $n'$, i.e., the average sizes of $L_u$ and $L_v$ are $n'$. The average time complexity of explaining a single test sample is $O(n'k)$ to compute inner products between normalized embeddings of size $k$.

\newpage
\section{Comparison to TracIn~\citep{pruthi2020estimating}}
\label{sec:compare_tracin}
TracIn traces the loss change of a given test point during training.
When the model being explained is trained with stochastic gradient descent, the loss change at each iteration can be attributed to a single training data we used to update the model. However, it is intractable since we do not know the test sample during training. \citet{pruthi2020estimating} proposes TracInCP as a practical alternative by measuring gradients similarities only on checkpoints $|\calT|$ of the model. TracInCP has the following formulation: 
\begin{equation}
\label{eqn:tracIn}
\mathbf{I}_{\mbox{TracInCP}}(z_i, z') = \sum_{t \in \calT} \eta^{(t)}
\nabla_{\theta} \calL(z_i, \theta^{(t)})^\top \nabla_{\theta} \calL(z', \theta^{(t)}),
\end{equation}
where $\eta^{(t)}, \theta^{(t)}$ are the learning rate and model parameters at iteration $t$, and $\calL$ is a loss function. 

To compare the TracIn with the high-dimensional representer, we perform experiments on collaborative filtering since $\ell_1$ regularized linear models in Section \ref{sec:exp_glm} are usually not trained with stochastic gradient descent. The experimental settings are the same as in Section \ref{sec:exp_cf}.
\begin{table*}[h!]
\small
\center
\begin{tabular}{|c|c|c|c|c|c|} 
\toprule
\multirow{2}*{Datasets} & \multirow{2}*{Models} & \multirow{2}*{Metrics} & 
\multicolumn{3}{c|}{Methods} \\
\cmidrule{4-6}
&&&  High-dim  Rep. & TracInCP & Random \\
\midrule
\multirow{4}*{\begin{tabular}{c}MovieLens-\\1M\end{tabular}} 
& \multirow{2}*{MF} 
& $\text{AUC-DEL}_{+} $ &  $-0.196  \pm 0.006$ & $\mathbf{-0.217} \pm 0.007$
& $-0.002 \pm 0.002$   \\
&& $\text{AUC-DEL}_{-} $ & $\mathbf{0.169}  \pm 0.004$  & $0.161 \pm 0.005$
& $-0.001 \pm 0.002$  \\
\cmidrule{2-6}
& \multirow{2}*{\begin{tabular}{c}Youtube-\\Net \end{tabular}} 
& $\text{AUC-DEL}_{+} $ &  $\mathbf{-0.227}  \pm 0.008$ &  $-0.161 \pm 0.008$
& $-0.001 \pm 0.004$   \\
&& $\text{AUC-DEL}_{-} $ & $\mathbf{0.214}  \pm 0.007$ & $0.165 \pm 0.007$
& $0.006 \pm 0.004$  \\
\midrule
\multirow{4}*{\begin{tabular}{c}
Amazon \\ reviews 2018  \\ (video games) 
\end{tabular}} 
& \multirow{2}*{MF} 
& $\text{AUC-DEL}_{+} $ &  $-0.184  \pm 0.012$ & $\mathbf{-0.312} \pm 0.013$
& $-0.070 \pm 0.011$   \\
&& $\text{AUC-DEL}_{-} $ & $0.080 \pm 0.012$ & $\mathbf{0.158} \pm 0.012$
& $-0.077 \pm 0.011$  \\
\cmidrule{2-6}
& \multirow{2}*{\begin{tabular}{c}Youtube-\\Net \end{tabular}} 
& $\text{AUC-DEL}_{+} $ &  $-0.234 \pm 0.014$ &  $\mathbf{-0.245} \pm 0.016$
& $-0.032 \pm 0.011$   \\
&& $\text{AUC-DEL}_{-} $ & $\mathbf{0.294}  \pm 0.011$  & $0.276 \pm 0.011$
& $-0.032 \pm 0.011$  \\
\bottomrule
\end{tabular}
\caption{Case deletion diagnostics for removing positive (negative) impact training samples on various datasets and models. $95\%$ confidence interval of averaged deletion diagnostics on $40 \times 40 = 1,600$ samples is reported.
Smaller (larger) $\text{AUC-DEL}_{+}$ ($\text{AUC-DEL}_{-}$) is better.}
\label{table:exp_tracIn}
\end{table*}

Table \ref{table:exp_tracIn} shows the results. We observe that although in general TracIn is worse than the high-dimensional representer on YoutubeNet, the performance of tracIn is much better than the high-dimensional representers on vanilla MF models. We argue that TracIn for MF has the same formulation as the high-dimensional representers on any checkpoints, so it can be viewed as an ensemble of high-dimensional representers over trajectories. 

By applying TracInCP formula (Eqn.\eqref{eqn:tracIn}) to the MF objective (Eqn.\eqref{eqn:general_mf}), the TracInCP formulation is as below:
given a training point $(i, j) \in \calD$, and a test sample $(i', j')$, the TracInCP importance scores on MF models are
\begin{equation}
\mathbf{I}_{\mbox{TracInCP}}((i,j), (i',j')) =  
\begin{cases}
-\sum_{t \in \calT} \eta^{(t)} \ell'_{\theta^{(t)}} ( y_{ij}, \langle \tilde{U}_i, \tilde{V}_{j} \rangle ) \ \langle \tilde{U}_i, \tilde{U}_{i'} \rangle
& \text{if $j = j'$.} \\
-\sum_{t \in \calT} \eta^{(t)}\ell'_{\theta^{(t)}}( y_{ij}, \langle \tilde{U}_i, \tilde{V}_{j} \rangle  )\ \langle \tilde{V}_j, \tilde{V}_{j'} \rangle  & \text{if $i = i'$.}  \\
0 & \text{ otherwise.}\\
\end{cases}
\end{equation}
Therefore, it has the same formulation as the high-dimensional representer for CF in Definition~\ref{def:L1_representers} except that TracInCP ensembles over the checkpoints on the trajectories. We note that Eqn.\eqref{eqn:tracIn} uses the same loss function $\calL$ for both training and test samples. However, our evaluation criteria is to find positive (or negative) impact samples, which is not necessarily the triaining loss. Therefore, we replace the test loss $\calL$ in Eqn.\eqref{eqn:tracIn} with prediction score $\langle U_{i'}, V_{j'}\rangle$ and then calculate its gradients.

\newpage
\section{Omitted Proofs}
\label{sec:proof}

\subsection{Proof of Theorem \ref{thm:rep_unified_reg}}
\begin{proof}
Since $\hat{\theta}$ is the minimizer of Eqn.\eqref{eqn:obj_unified_reg}, there exists a subgradient of the regularization $r(\cdot)$, $g = u_\theta + v$ such that 
$$
0 = \frac{\partial}{\partial \theta} \left(\frac{1}{n} \sum_{i=1}^n \ell(y_i, \langle x_i, \hat{\theta} \rangle) + \lambda r(\theta) \right)
= \frac{\partial}{\partial \theta} \left(\frac{1}{n} \sum_{i=1}^n \ell(y_i, \langle x_i, \hat{\theta} \rangle) \right) + \lambda g.$$
It implies
$$
g = - \frac{1}{n\lambda} \sum_{i=1}^n \ell'(y_i, \langle x_i , \hat{\theta} \rangle) x_i.
$$
After applying inverse transform $(\partial_{\theta} r)^+$ of the partial differential on both sides, we have 
\begin{align*}
\hat{\theta} =  (\partial_{\theta} r)^+ 
(g) = - \frac{1}{n\lambda} \sum_{i=1}^n \ell'(y_i, \langle x_i , \hat{\theta} \rangle) \left(\partial_{\theta} r)^+x_i\right) .
\end{align*}
By taking inner product with $x'$ on both sides, we have 
$$
\langle x', \hat{\theta} \rangle = \sum_{i=1}^n  - \frac{1}{n\lambda}   \ell'(y_i, \langle x_i, \hat{\theta} \rangle))  
\langle  (\partial_{\hat{\theta}} r)^{\frac{+}{2}} x_i ,(\partial_{\hat{\theta}} r)^{\frac{+}{2}}x' \rangle.
$$
\end{proof}

\subsection{Proof of Corollary \ref{thm:rep_l1_reg}}
\begin{proof}

Below is another proof without applying Theorem \ref{thm:rep_unified_reg}:

Since $\hat{\theta}$ is the minimizer of Eqn.\eqref{eqn:obj_l1_reg}, there exists a subgradient of the $\ell_1$ norm, $v = \partial \norm{\hat{\theta}}{1} \in \real^p$ such that 
$$
0 = \frac{\partial}{\partial \theta} \left(\frac{1}{n} \sum_{i=1}^n \ell(y_i, \langle x_i, \hat{\theta} \rangle) + \lambda \norm{\hat{\theta}}{1} \right)
= \frac{\partial}{\partial \theta} \left(\frac{1}{n} \sum_{i=1}^n \ell(y_i, \langle x_i, \hat{\theta} \rangle) \right)+ \lambda v,$$
where the $i^{th}$ coordinate of $v$ is $v_i = \mbox{sign}(\theta_i)$ if $\hat{\theta}_i \neq 0$, and $-1 \leq v_i \leq 1$ if $\hat{\theta}_i = 0$ for all $i \in [p]$. The above equation implies:
$$
v = - \frac{1}{n\lambda} \sum_{i=1}^n \ell'(y_i, \langle x_i , \hat{\theta} \rangle) x_i.
$$
Since we have $\hat{\theta} = |\hat{\theta}| \odot v$, by coordinate-wisely multiplying $|\hat{\theta}|$ on both sides, we have 
\begin{align*}
\hat{\theta}  = |\hat{\theta}| \odot v = - \frac{1}{n\lambda} \sum_{i=1}^n \ell'(y_i, \langle x_i , \hat{\theta} \rangle) ( |\hat{\theta}| \odot x_i ).
\end{align*}
By taking an inner product with $x'$ on both sides, we have 
$$
\langle x', \hat{\theta} \rangle = \sum_{i=1}^n  - \frac{1}{n\lambda}   \ell'(y_i, \langle x_i, \hat{\theta} \rangle))  
\langle  \sqrt{|\hat{\theta}|} \odot x_i  , \sqrt{|\hat{\theta}|} \odot x' \rangle.
$$
\end{proof}

\subsection{Proof of Corollary\ref{thm:rep_nuc_reg}}

\begin{proof}
According to page 40 in~\citet{watson1992characterization}, the subgradient of nuclear norm at $\hat{\Theta}$ is 
$$
\partial \norm{\hat{\Theta}}{*} = \{ 
UV^{\top} + W : W \in \real^{d_1 \times d_2}, \norm{W}{2} \leq 1, W V = \mathbf{0}, U^{\top}W = \mathbf{0}
\}.
$$
Since $\hat{\Theta}$ is the minmizer of Eqn.\eqref{eqn:obj_nuclear_norm_reg}, we can find a subgradient $ UV^T + W$ such that the derivative of Eqn.\eqref{eqn:obj_nuclear_norm_reg} with respect to $\Theta$ at $\hat{\Theta}$ is zero:
\begin{align*}
& 0 = \frac{1}{n} \sum_{i=1}^n \ell'(y_i, \langle X_i , \hat{\Theta} \rangle) X_i + \lambda ( UV^{\top} - W) \\
& \Rightarrow 
UV^{\top} = - \frac{1}{n\lambda} \sum_{i=1}^n \ell'(y_i, \langle X_i , \hat{\Theta} \rangle) X_i - W
\end{align*}
Next, by using the fact that $U^{\top}U = I_k$ is a identity matrix, we have 
\begin{align*}
\hat{\Theta} 
& = U \Sigma V^{\top} = U\Sigma  U^{\top}U  V^{\top} \\
& = U\Sigma U^{\top} \left( - \frac{1}{n\lambda} \sum_{i=1}^n \ell'(y_i, \langle X_i , \hat{\Theta} \rangle) X_i - W \right) \\
& =  \sum_{i=1}^n \left( - \frac{1}{n\lambda} \ell'(y_i, \langle X_i , \hat{\Theta} \rangle) \right)
\left(  U\Sigma U^{\top} X_i  \right) \ \ \  \text{(Using the fact that } U^{\top}W = \mathbf{0} ) 
\end{align*}
Similarly, we have
$$
\hat{\Theta} 
= \sum_{i=1}^n - \frac{1}{n\lambda}   \ell'(y_i, \langle X_i, \hat{\Theta} \rangle_F)
\left(   X_i V^{\top}\Sigma V \right).
$$
By taking inner product w.r.t a test sample $X'$, we have  
\begin{align*}
\langle X', \hat{\Theta} \rangle_F
& = \sum_{i=1}^n  - \frac{1}{n\lambda}   \ell'(y_i, \langle X_i, \hat{\Theta} \rangle))  
\langle  \sqrt{\Sigma} U^{\top} X_i  , \sqrt{\Sigma} U^{\top} X' \rangle_F \\ 
& = \sum_{i=1}^n  - \frac{1}{n\lambda}   \ell'(y_i, \langle X_i, \hat{\Theta} \rangle))  
\langle   X_i V \sqrt{\Sigma} , X' V \sqrt{\Sigma}  \rangle_F,
\end{align*}
using the fact that $\langle A, B \rangle_F = \trace{A^{\top}B}$ and the cyclic property of the trace operator.
\end{proof}

\subsection{Proof of Corollary~\ref{cor:rep_mf}}
\begin{proof}
We first show that the optimization problem in Eqn.\eqref{eqn:mf_with_nuc_reg} is a special case of Eqn.\eqref{eqn:obj_nuclear_norm_reg}. Denote the $k^{th}$ data in $\calD$ as $(i_k,j_k)$. Let $X_k \in \real^{|\calU| \times |\calI|}$ a zero matrix except that the $(i_k,j_k)$ coordinate is $1$, and  $y_k = Y_{i_k, j_k}$. By plugging the $X_k, y_k$ to Eqn.\eqref{eqn:obj_nuclear_norm_reg} for $1 \leq k \leq |\calD|$, we recover Eqn.\eqref{eqn:mf_with_nuc_reg}.

Let $X' \in \real^{|\calU| \times |\calI|}$ be a zero matrix except that the $(i',j')$ coordinate is $1$. 
By applying Corollary \ref{thm:rep_nuc_reg}, we have 
\begin{align*}
\langle X', \hat{\Theta}
\rangle 
& = \hat{\Theta}_{i',j'} \\
& = \sum_{k=1}^{|\calD|}  - \frac{1}{\lambda|\calD|}   \ell'(y_k, \langle X_k, \hat{\Theta} \rangle))  
\langle  \sqrt{\Sigma} U^{\top} X_k  , \sqrt{\Sigma} U^{\top} X' \rangle_F \\
& = \sum_{k=1}^{|\calD|} - \frac{1}{\lambda|\calD|}   \ell'(y_k, \hat{\Theta}_{i_k, j_k})) \trace{\sqrt{\Sigma}U^{\top}X'X_k^{\top} U\sqrt{\Sigma} }. \\
\end{align*}
Since 
$X'X_k$ is a zero matrix if $j_k \neq j'$ and is a zero matrix except that the entry $(i_k,i')$ is one if $j_k = j'$, we have 
\begin{align*}
\hat{\Theta}_{i',j'}
& = \sum_{k=1}^{|\calD|} - \frac{1}{\lambda|\calD|}   \ell'(y_k, \hat{\Theta}_{i_k, j_k})) \indic{j_k=j'}
\langle \sqrt{\Sigma}U_{i'}, \sqrt{\Sigma}U_{i_k} \rangle \\
& = \sum_{i:(i,j') \in \calD} - \frac{1}{\lambda|\calD|} \ell'( y_{ij}, \hat{\Theta}_{ij}) \langle \sqrt{\Sigma}U_i, \sqrt{\Sigma}U_{i'}\rangle.\\
\end{align*}
Similarly, by using Eqn.\eqref{eqn:nuc_reg_decomp2}, we have 
$$
\hat{\Theta}_{i',j'} = \sum_{j:(i',j)\in \calD} - \frac{1}{\lambda|\calD|} \ell'( y_{ij}, \hat{\Theta}_{ij}) \langle \sqrt{\Sigma}V_j, \sqrt{\Sigma}V_{j'} \rangle.
$$
\end{proof}

\end{document}